\documentclass[times,twocolumn,final]{elsarticle}

\usepackage{ipcai_arxiv}
\usepackage{framed,multirow}

\usepackage{amssymb}
\usepackage{latexsym}

\usepackage{url}
\usepackage{cite}
\usepackage{amsmath,amssymb,amsfonts,bm}
\usepackage{booktabs}
\usepackage{arydshln}
\usepackage{tabulary}
\usepackage{graphicx,epstopdf}
\usepackage{textcomp}
\usepackage{subcaption}
\usepackage{booktabs}
\usepackage[svgnames,table]{xcolor}
\usepackage{pifont}
\usepackage{arydshln}
\usepackage[colorlinks=true,citecolor=cyan,urlcolor=cyan]{hyperref}
\usepackage{graphicx}
\usepackage{placeins}
\usepackage{fancyhdr}
\usepackage{float}

\usepackage{listings}%
\usepackage{lipsum}
\usepackage{makecell}

\fancypagestyle{firstpagestyle}{
    \fancyhf{} 
    \fancyhead{} 
    \fancyfoot{} 
    \fancyhead[CO]{\em \fontsize{9pt}{8pt}\selectfont This article has been accepted to the International Conference on Information Processing in Computer-Assisted Interventions, 2026.}
}

\begin{document}


\title{Where are they looking in the operating room?}

\author[1]{Keqi \snm{Chen}\corref{cor}\fnref{co-first}}

\author[1]{Séraphin \snm{Baributsa}\fnref{co-first}}

\author[3,5]{Lilien \snm{Schewski}\fnref{co-first}}

\author[1,2]{Vinkle \snm{Srivastav}}

\author[2,4]{Didier \snm{Mutter}}

\author[3,5]{Guido \snm{Beldi}}

\author[3,5]{Sandra \snm{Keller}}

\author[1,2]{Nicolas \snm{Padoy}}

\cortext[cor]{Corresponding author: keqi.chen@unistra.fr}
\fntext[co-first]{shared first authorship}

\address[1]{University of Strasbourg, CNRS, INSERM, ICube, UMR7357, France}

\address[2]{IHU Strasbourg, 67000 Strasbourg, France}

\address[3]{Department for Biomedical Research (DBMR), University of Bern, 3008 Bern, Switzerland}

\address[4]{University Hospital of Strasbourg, 67000 Strasbourg, France}

\address[5]{Department for Visceral Surgery and Medicine, Bern University Hospital, University of Bern, 3010 Bern, Switzerland}

\received{XXX}
\finalform{XXX}
\accepted{XXX}
\availableonline{XXX}
\communicated{XXX}

\begin{abstract}

\textbf{Purpose: } 
Gaze-following, the task of inferring where individuals are looking, has been widely studied in computer vision, advancing research in visual attention modeling, social scene understanding, and human–robot interaction. However, gaze-following has never been explored in the operating room (OR), a complex, high-stakes environment where visual attention plays an important role in surgical workflow analysis. In this work, we introduce the concept of gaze-following to the surgical domain, and demonstrate its great potential for understanding clinical roles, surgical phases, and team communications in the OR. 

\textbf{Methods: } 
We extend the 4D-OR dataset with gaze-following annotations, and extend the Team-OR dataset with gaze-following and a new team communication activity annotations. Then, we propose novel approaches to address clinical role prediction, surgical phase recognition, and team communication detection using a gaze-following model. For role and phase recognition, we propose a gaze heatmap-based approach that uses gaze predictions solely; for team communication detection, we train a spatial-temporal model in a self-supervised way that encodes gaze-based clip features, and then feed the features into a temporal activity detection model. 

\textbf{Results: } 
Experimental results on the 4D-OR and Team-OR datasets demonstrate that our approach achieves state-of-the-art performance on all downstream tasks. Quantitatively, our approach obtains F1 scores of 0.92 for clinical role prediction and 0.95 for surgical phase recognition. Furthermore, it significantly outperforms existing baselines in team communication detection, improving previous best performances by over 30\%. 

\textbf{Conclusion: } 
We introduce gaze-following in the OR as a novel research direction in surgical data science, highlighting its great potential to advance surgical workflow analysis in computer-assisted interventions. Although limited to monocular 2D gaze prediction relying on manual annotations, our research clearly demonstrates the clinical value of gaze analysis from ceiling-mounted cameras. Future work will explore semantic understanding, multi-view learning, and few-shot approaches to further improve scalability and robustness.
\end{abstract}

\maketitle
\thispagestyle{firstpagestyle}

\begin{figure*}[h!]
    \centering
    \includegraphics[width=0.8\textwidth]{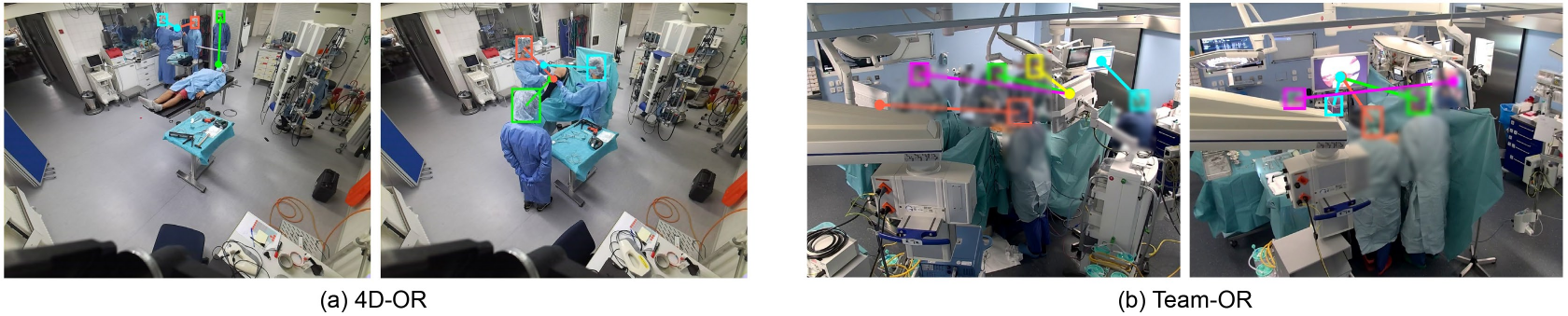} 
    \caption{Examples of the gaze-following annotations on the 4D-OR~\citep{ozsoy20224d} and Team-OR~\citep{chen2025they} datasets, where bounding boxes represent heads and arrows denote the corresponding gaze points.}
    \label{fig:gaze_annotations}
\end{figure*}

\section{Introduction}\label{sec:intro}

Gaze is a powerful indicator of human attention, intention, and interaction~\citep{emery2000eyes}. In the highly complex modern operating room (OR), understanding OR team members' visual attention through gaze analysis offers a unique perspective on holistic surgical scene analysis. Given the diverse roles and responsibilities within the OR teams, including surgeons, nurses, and anesthesiologists, each team member exhibits specific gaze behavior patterns that reflect their specific tasks and attentional demands during different phases of the operation. For instance, surgeons primarily focus on the operative field, anesthesiologists monitor patient vital signs and anesthesia equipment, while nurses alternate attention between instrument management, environmental monitoring, and coordination tasks. Beyond these individual attentional differences, gaze also serves as a crucial channel for non-verbal communication in the OR. Through mutual eye contact, shared visual focus, or referential gaze (where one agent tries to induce another agent’s attention to a target via gaze)~\citep{fan2019understanding}, team members exchange critical information, maintain mutual awareness, and coordinate their actions efficiently.

Over the past years, the surgical data science community has increasingly recognized the value of gaze information, and has conducted extensive research using wearable or head-mounted eye-tracking systems to capture and analyze OR team members’ visual attention during both simulated~\citep{james2007eye,zheng2011surgeon,bhavsar2017quantifying,black2018auditory,reed2024evaluation,ozsoy2025egoexor} and real surgeries~\citep{seagull1999monitoring,khan2012analysis,atkins2013surgeons,erridge2018comparison,reed2024evaluation}. However, implementing such systems in real surgical procedures remains challenging. Clinicians are often reluctant to use additional wearable equipment, because the continuous wearing throughout the operation can interfere with their comfort, mobility, and concentration, disturbing the normal workflow of the surgery. Furthermore, configuring multiple eye trackers to obtain synchronized, ego-centric gaze data from several team members substantially increases technical difficulty, as it often requires using visual markers~\citep{niehorster2025gazemapper}, physical actions~\citep{burger2018synchronizing}, hardware triggers~\citep{atkins2013surgeons}, or specialized software protocols~\citep{seagull1999monitoring,ozsoy2025egoexor}. These challenges make large-scale deployment difficult to achieve.

To overcome these limitations, an alternative approach is to predict where clinicians are looking by following their gaze directions in a holistic view (e.g. ceiling-mounted cameras), which is often referred to as gaze-following~\citep{recasens2015they}. Unlike traditional eye-tracking systems, gaze-following does not require additional devices, and has been extensively studied in the general computer vision field~\citep{recasens2015they,recasens2017following,chong2018connecting,chong2020detecting,tu2022end,tonini2023object,ryan2025gaze}, along with several downstream applications such as intention prediction~\citep{wei2018and}, human-object interaction detection~\citep{xu2019interact}, gaze communication detection~\citep{fan2018inferring,fan2019understanding}, and human-robot interaction~\citep{saran2018human,jin2022depth}. However, gaze-following and its applications in the OR still remain unexplored. 

In this work, we aim to introduce gaze-following research into the surgical domain. Specifically, we extend the 4D-OR~\citep{ozsoy20224d} and Team-OR~\citep{chen2025they} datasets with gaze annotations, including head bounding boxes and corresponding gaze points if the OR team members are looking inside the view. Then, we train a foundation model-based gaze estimation model~\citep{ryan2025gaze} that can accurately follow the OR team members' gazes in the complex OR scenes. Afterwards, we propose novel approaches to address three downstream tasks using the trained gaze-following model, including clinical role prediction and surgical phase recognition on the 4D-OR dataset, and team communication detection on the Team-OR dataset. For role and phase recognition, we propose a transformer-based approach that processes generated gaze heatmaps, and evaluate the classification results on the 4D-OR dataset using existing annotations. 

For team communication detection, we propose to detect the ``StOP?'' activities~\citep{chen2025they}, where all OR team members pause their work and communicate for about 30-90 seconds during the surgical procedure. Furthermore, we extend the Team-OR dataset with a more fine-grained new activity class named ``anesthesiologists being attentive (A.B.A.)'', where we annotate the moments when anesthesiologists concentrate on observing the surgical field or the laparoscopic monitor of the surgeons performing operational procedures. Such observable behaviors are theoretically linked to the concept of situation awareness, which enables team members to perceive elements in a situation, comprehend their meaning, and forecast future situation events and dynamics~\citep{gaba1995situation,endsley2000theoretical}. 
In order to detect both ``StOP?'' and ``A.B.A.'' activities in the untrimmed surgical videos, we propose a self-supervised learning strategy, where we train a spatial-temporal transformer that encodes clip-level gaze features under supervision of the off-the-shelf visual features~\citep{wang2023videomae}. Then, we use gated fusion to augment video features with gaze features, and pass them through a temporal activity detection (TAD) network to obtain the results. Experiments on the 4D-OR and Team-OR datasets prove the value of studying gaze-following in the OR. 

Our contributions can be summarized as follows: (1) we highlight the clinical importance of studying gaze-following in the OR for surgical workflow analysis, and extend two existing external OR datasets with gaze annotations; (2) we present a transformer-based approach that processes generated gaze heatmaps solely to address clinical role prediction and surgical phase recognition on the 4D-OR dataset, and reach a performance on par with the semantic scene graph-based approaches; (3) we extend the Team-OR dataset with a new activity label, propose a novel self-supervised learning strategy to train a spatial-temporal model that encodes clip-level gaze features, and achieve state-of-the-art performance on the team communication detection task. 




\begin{figure*}[h]
\centering
\includegraphics[width=0.75\textwidth]{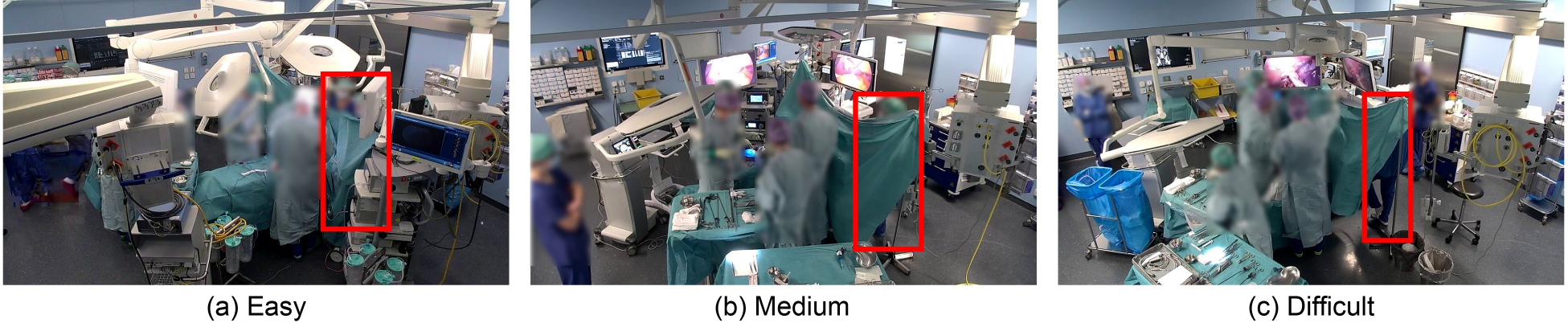}
\caption{Examples of the ``anesthesiologists being attentive'' moments from easy to difficult: (a) two anesthesiologists are looking at the operating table; (b) one is looking at the monitor screen, which is close to the anesthesia equipment; (c) one is observing the surgery but heavily occluded. }\label{fig:ana}
\end{figure*}

\section{Methodology}\label{sec:method}
\subsection{Gaze-following datasets in the operating room}\label{sec:gaze-following}

To enable gaze-following study in the OR, we extend the 4D-OR~\citep{ozsoy20224d} and Team-OR datasets~\citep{chen2025they} with gaze annotations. 4D-OR is a video dataset consisting of 10 simulated knee surgeries with annotations of four clinical roles (head surgeon, assistant surgeon, circulating nurse and anesthetist) and eight surgical phases (OR preparation, patient roll-in, patient preparation, implant placement preparation, implant placement, conclusion, patient roll-out, and OR cleanup). The video recording is done in one frame per second, resulting in a total of 6,734 frames. The Team-OR dataset consists of more than 100 hours of operation videos of 37 real laparoscopic surgeries with team communication annotations (e.g. ``StOP?'' activity, where all OR team members pause their work and communicate for about 30-90 seconds during the surgical procedure). We sample one frame every five minutes, resulting in a total of 1,284 frames. For both datasets, we select one view that mostly captures the holistic OR scenes with the largest number of persons visible for annotations. 

After preparing the images, we conduct gaze-following annotations following the proposed protocol in previous works~\citep{recasens2015they,chong2020detecting}. Specifically, 
we first annotate the head bounding boxes for each person in each image. Then, we label the gaze target of each annotated person as a single gaze point based on where we believe the person is looking. We also mark if the person is looking outside the image. To ensure the quality, we have two reviewers who double-check the annotations. In the end, we have gaze annotations of 18,743 and 5,870 persons in the 4D-OR and Team-OR datasets, respectively. Examples of the gaze annotations are shown in Fig.~\ref{fig:gaze_annotations}.


\subsection{Extended Team-OR dataset}

In addition to gaze annotations, we extend the Team-OR dataset~\citep{chen2025they} with a new activity class named ``anesthesiologists being attentive (A.B.A.)'' using MOSaiC platform~\citep{mazellier2023mosaic}. Unlike the existing ``StOP?'' activity that concerns all OR team members, ``A.B.A.'' is a more fine-grained activity that centers on the anesthesiologists. Specifically, it is characterized by anesthesiologists moving toward the operating table and concentrating on observing the surgical field or the laparoscopic monitor of the surgeons performing operational procedures, as shown in Fig.~\ref{fig:ana}. After annotating all the videos, we have 1,466 ``A.B.A.'' activities in total, with an average duration of 24 seconds. Since the definition of ``A.B.A.'' activity is highly correlated with human gaze, we propose to evaluate the value of studying gaze-following by detecting these moments in the untrimmed videos. The challenges are three-fold: (1) the dramatic duration variance, as the longest one lasts 366 seconds, while the shortest one only lasts 1 second; (2) the ambiguity of similar activities, such as adjusting anesthesia equipment (which is close to laparoscopic monitor) or interacting with a colleague; (3) the occlusions by colleagues or medical equipment, as shown in Fig.~\ref{fig:ana}.

\subsection{Clinical role prediction and surgical phase recognition}

\begin{figure*}[h]
\centering
\includegraphics[width=0.9\textwidth]{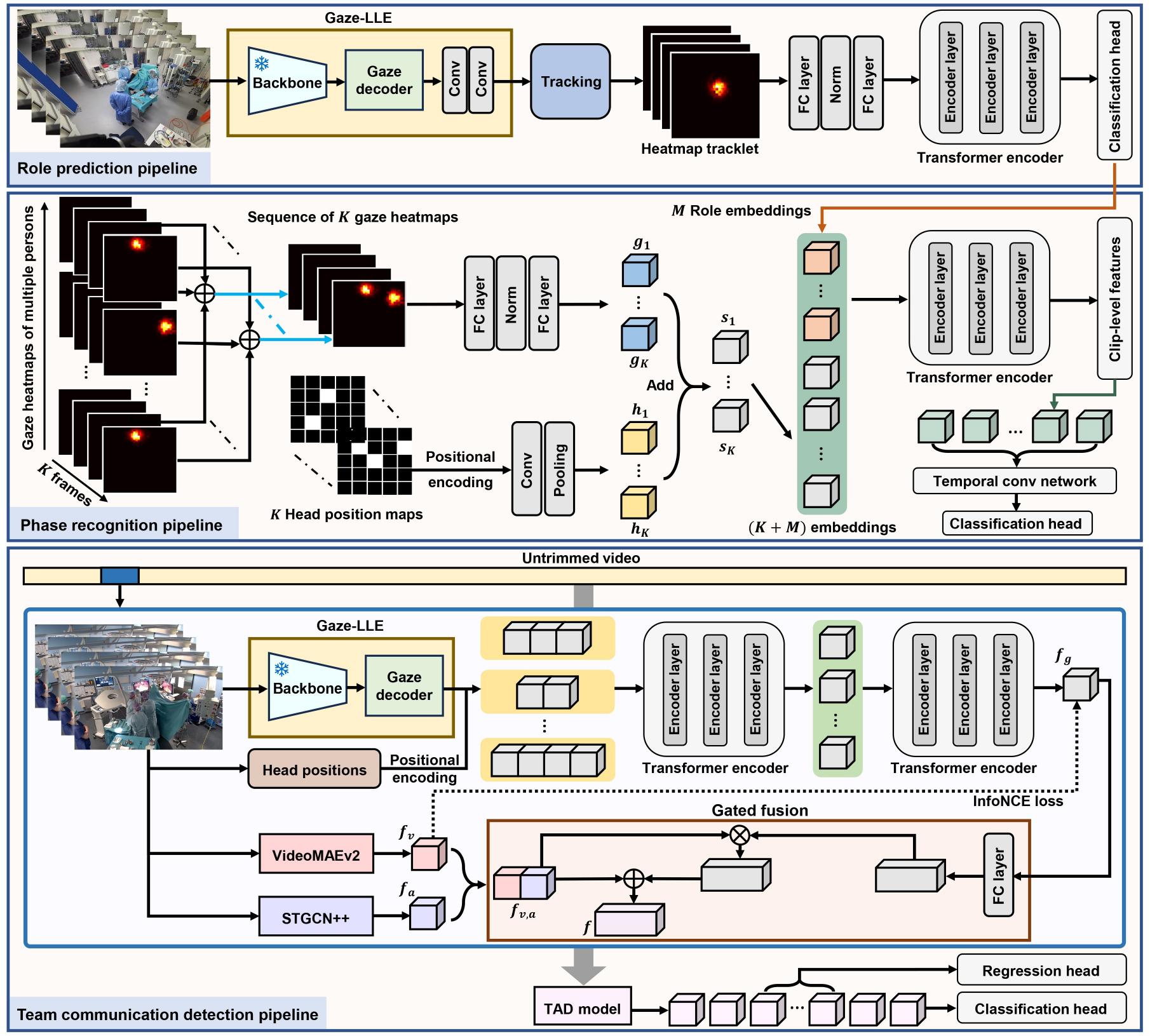}
\caption{Pipelines of our methods for the downstream tasks (FC = Fully-Connected, TAD = Temporal Activity Detection). (1) Role prediction: we use Gaze-LLE~\citep{ryan2025gaze} and tracking to generate heatmap tracklets, and pass them through a transformer encoder to classify the roles. (2) Phase recognition: we fuse features of gaze heatmaps and head positions of each frame, pass sequential frame-level features through a transformer encoder along with learnable role embeddings to obtain clip-level features, and pass all the clip-level features through a temporal convolution network to classify the surgical phases. (3) Team communication detection: we train a spatial-temporal transformer that encodes gaze features for each clip through contrastive learning with visual features~\citep{wang2023videomae}, merge gaze, visual and action features~\citep{duan2022pyskl} through gated fusion, and then detect the activities through a TAD model~\citep{chen2025they}.} \label{fig:pipelines}
\end{figure*}

With gaze annotations, we fine-tune a state-of-the-art gaze-following model~\citep{ryan2025gaze} using OR data, whose backbone is frozen during training. The model generates a gaze heatmap $\mathcal{H} \in \mathcal{R}^{H \times W}$ for each person, and thus we use $\mathcal{H}$ as the sole input for clinical role prediction and surgical phase recognition. The frameworks are shown in Fig.~\ref{fig:pipelines}. 

To predict each person's role, we follow the tracking strategy in \citep{ozsoy20224d} to track each person, use Hungarian matching to associate heads and full bodies, and thus obtain a gaze heatmap tracklet $\mathcal{T} \in \mathcal{R}^{T \times H \times W}$ for each person. Then, we pass $\mathcal{T}$ through linear layers and a three-layer transformer encoder and compute the probability scores for each role. We train the model using cross-entropy loss. During inference, we further apply the unique role assignment~\citep{ozsoy20224d}, which guarantees that each human in a frame is assigned to a distinct role.

For phase recognition, we encode gaze heatmaps within the nearest $K$ frames to classify the central frame. For each frame $i$, we first sum up the normalized heatmaps into one, and pass it through fully-connected layers to obtain a frame-level gaze embedding $g_i$. Then, we generate a sparse 2D head position map for each frame $i$ through positional encoding~\citep{carion2020end}, and pass the map through convolution and adaptive pooling layers to obtain a frame-level head embedding $h_i$. We add both embeddings to compute the frame-level scene embedding $s_i$: 
\begin{equation}
    s_i = g_i + h_i.
    \label{eq:scene_emb}
\end{equation}

In our model, we also maintain learnable embeddings of $M$ roles. Based on the role prediction results, we prepare input role embeddings. If a role does not exist in the $K$ frames, we set its embedding to zero. Subsequently, we pass the obtained $K$ scene embeddings and $M$ role embeddings as $(K + M)$ tokens through a three-layer transformer encoder and obtain the clip-level features. Finally, we pass all the clip-level features through a temporal convolution network~\citep{farha2019ms} and classify the output. We use cross-entropy loss during training. 



\subsection{Team communication detection}

In this section, we combine the gaze-following model with the state-of-the-art TAD framework~\citep{chen2025they} to detect the ``StOP?'' and ``A.B.A.'' activities in real surgeries, as shown in Fig.~\ref{fig:pipelines}. Specifically, we propose a self-supervised learning strategy to train a spatial-temporal transformer encoder that encodes clip-level gaze features. 

For each video clip comprising $N$ frames, where each frame contains $P$ persons (with $P$ varying across frames), we extract gaze features of size $G$ for each person from the gaze decoder of the gaze-following model without generating gaze heatmap. This process yields a sequence of gaze feature vectors $\mathcal{C} \in \mathcal{R}^{N \times P \times G}$ over time. Subsequently, we apply the spatial-temporal transformer encoder: for each frame, we pass the gaze vectors $\mathcal{C}_s \in \mathcal{R}^{P \times G}$ with head positions encoded through a three-layer encoder to obtain a frame-level vector from the \texttt{[CLS]} token, and then we pass these frame vectors $\mathcal{C}_t \in \mathcal{R}^{N \times G}$ through another three-layer encoder to obtain the final clip-level features $f_g$ of size $V$ from the \texttt{[CLS]} token. 

Considering the sparsity and ambiguity of the TAD labels, we train the gaze encoder under supervision of the visual features in a self-supervised way. Specifically, we use an off-the-shelf VideoMAEv2 model~\citep{wang2023videomae} to obtain visual features $f_v$ of size $V$ for each clip. As $f_v$ contains rich semantics and serves as a strong baseline in the TAD task~\citep{chen2025they}, we propose to densely match the gaze features $f_g$ with $f_v$ to ensure semantic consistency. Concretely, we sample a batch of clips and obtain the gaze and visual features. Then, we use InfoNCE loss~\citep{oord2018representation} to conduct contrastive learning:
\begin{equation}
    \mathcal{L}_{\mathrm{InfoNCE}} = -\frac{1}{B} \sum_{i=1}^{B} \log \frac{\exp(\mathrm{sim}(f_g^i, f_v^i) / \tau)}{\sum_{j=1}^{B} \exp(\mathrm{sim}(f_{g}^i, f^{j}_{v}) / \tau)},
    \label{eq:infonce_single}
\end{equation}
where $B$ is the batch size, $\text{sim}(\cdot,\cdot)$ is the cosine similarity, and $\tau$ is the temperature. 

Afterwards, we use the gaze features to augment existing features in the TAD framework. For each clip at time $t$, we use a gated fusion strategy that translates the gaze features $f_g^t$ into importance scores through a fully-connected layer and the sigmoid function. Then, after obtaining the concatenated visual and action features~\citep{duan2022pyskl} $f_{v,a}^t$, we use dot product along with a residual connection to obtain the final clip features:
\begin{equation}
    f^{t} = f_{v,a}^{t} + \left( \sigma(\mathbf{W} \cdot f_{g}^{t} + b) \odot f_{v,a}^{t} \right),
    \label{eq:gated_fusion_merged}
\end{equation}
where $\sigma(\cdot)$ is the sigmoid function and $\mathbf{W}$ and $b$ are weights and biases. 

Finally, we pass all the clip features through the TAD model, obtain the instant features, and detect the activities using a regression head and a classification head. During training, we use the same regression and classification losses as in \citep{chen2025they}. During inference of detecting ``A.B.A.'', since it is an anesthesiologist-centric activity, we heuristically only encode the gazes of the rightmost three persons that are near the anesthesia equipment in each frame (see Sec.~\ref{sec:abl}). 

\begin{figure*}
    \centering
    \includegraphics[width=0.9\textwidth]{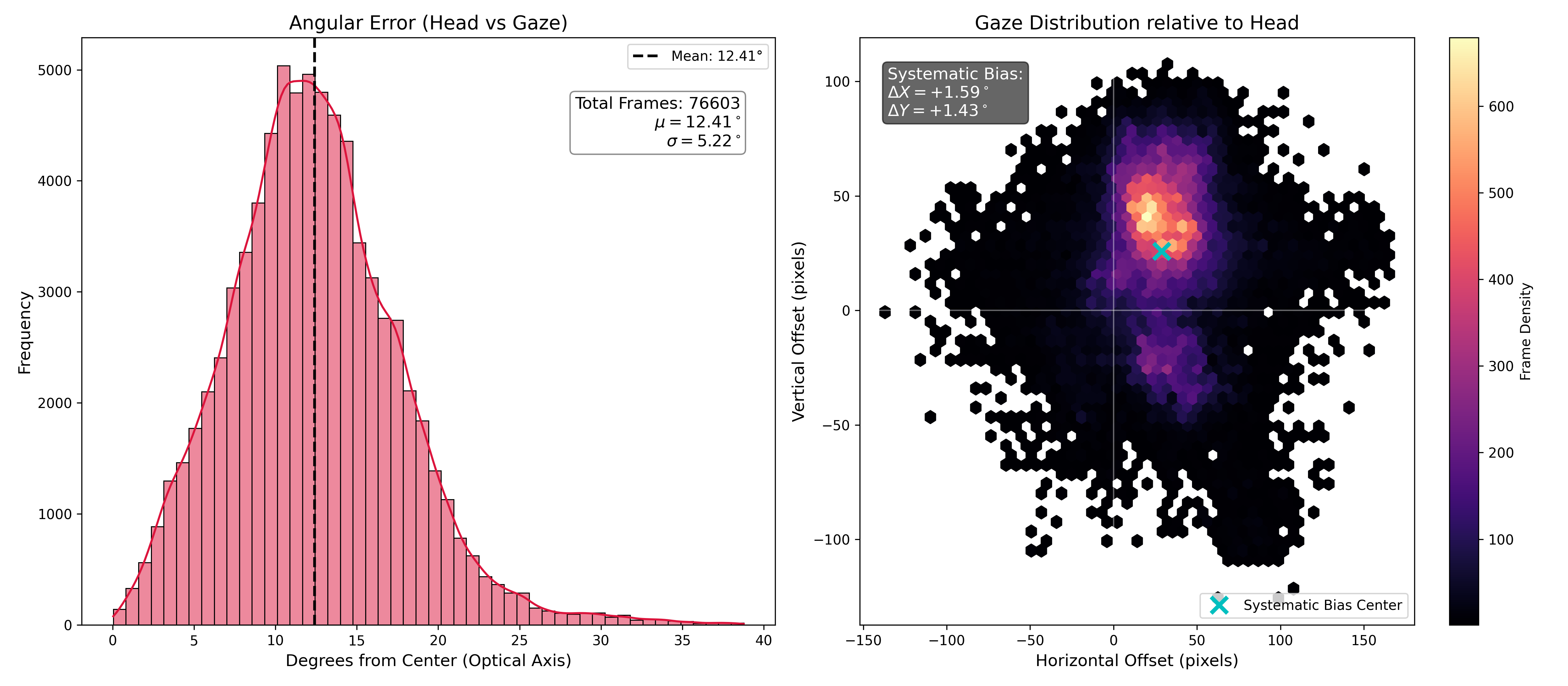} 
    \caption{Analysis of deviation between head orientation and gaze signal using the EgoExOR dataset~\citep{ozsoy2025egoexor} with 76,603 frames across multiple surgical procedures. (Left) Distribution of angular deviation between the head-worn camera's optical axis and gaze. (Right) 2D spatial distribution of gaze relative to the head center (frame center at (0,0)), where image resolution is 1408*1408.}
    \label{fig:sanity_check}
\end{figure*}

\section{Experiments}\label{sec:experiments}

\subsection{Gaze-following annotation validation}

In Sec.~\ref{sec:gaze-following}, we annotate gaze points for each person using only the ceiling camera view. Although the annotation strategy has been evaluated and adopted in natural image research, it is intrinsically inaccurate because the annotators mostly follow the head orientation to label the gaze points without the aid of accurate eye-tracking devices. It remains unclear how large the potential error could be in the OR scenes. To address this concern, we conduct a quantitative validation experiment using the EgoExOR dataset~\citep{ozsoy2025egoexor}, which provides gaze signals synchronized with egocentric videos across multiple surgical procedures, allowing for a direct comparison between the camera's optical axis (head orientation) and the actual gaze direction. Specifically, we quantify the deviation in two aspects: the absolute angular deviation between the head-worn camera's optical axis and gaze signal, and the 2D spatial distribution of gaze points relative to the head center. The results are shown in Fig.~\ref{fig:sanity_check}.

For the absolute angular deviation in Fig.~\ref{fig:sanity_check} (Left), we report a mean deviation of 12.41° with a standard deviation of 5.22°. The distribution shows a clear unimodal peak. For the 2D spatial distribution in Fig.~\ref{fig:sanity_check} (Right), we report a systematic bias of 1.59° and 1.43° in horizontal and vertical axes. According to the visualization, the clinician’s gaze is highly concentrated around the optical axis (the frame center), and the overall dispersion is limited.

Based on the experimental results above, we find that the deviation between the gaze signal and the head orientation is overall limited, indicating that head orientation is a reliable proxy for gaze direction in the OR context. Additionally, our gaze annotations do not solely depend on the head orientation, but also on the localization of the attentional gaze target within the area of interest. Based on the prior knowledge of where clinicians might be looking, the annotators are usually able to determine the semantic targets (e.g., operating table, screen, a clinician, etc.). Therefore, the gaze annotation strategy is generally reliable in the OR for surgical workflow analysis.

\subsection{Dataset and evaluation metrics}

We evaluate gaze-following task on the 4D-OR~\citep{ozsoy20224d} and Team-OR~\citep{chen2025they} datasets. For 4D-OR, following the original configuration, we use 6, 2 and 2 videos for training, validation, and testing, respectively. For Team-OR, we use 80\% of videos for training and the rest for testing. We use the same evaluation metrics as in \citep{ryan2025gaze}: the heatmap area under the curve (AUC), which considers each heatmap pixel as a confidence score for the receiver operating characteristic (ROC) curve, and pixel L2, defined as the normalized Euclidean distance between the predicted and ground truth gaze points. 

For clinical role prediction and surgical phase recognition, we evaluate them on the 4D-OR dataset, and use F1 score as the metric. For team communication detection, we evaluate it on the Team-OR dataset, and split the dataset into a train set and a test set with 60\% and 40\% videos, respectively. We report the average precision (AP) at different temporal intersections over union (tIoU) thresholds as the metrics. 

\begin{table*}[h]
\caption{Comparison with the state-of-the-art temporal action detection approaches. We report AP at different tIoU. }\label{tab:team_results}
\centering
\resizebox{0.8\textwidth}{!}{
\begin{tabular*}{\textwidth}{@{\extracolsep\fill}c|cccccc}
\toprule%
\multirow{2}{*}{Method} & \multicolumn{6}{@{}c@{}}{StOP?} \\\cmidrule{2-7}%
 & 0.1 & 0.2 & 0.3 & 0.4 & 0.5 & Avg. \\
\midrule
ActionFormer~\citep{zhang2022actionformer}  & 12.63 & 12.63 & 12.63 & 5.79 & 1.22 & 8.98 \\
TriDet~\citep{shi2023tridet}  & 23.46 & 13.37 & 13.35 & 13.27 & 1.87 & 13.06 \\
TemporalMaxer~\citep{tang2023temporalmaxer} & 16.04 & 14.36 & 13.72 & 13.53 & 12.38 & 14.01 \\
Team-OR~\citep{chen2025they}  & \textbf{30.85} & 20.78 & 20.66 & 20.58 & 20.52 & 22.68 \\
Ours (off-the-shelf) & 25.42 & 25.40 & 25.30 & 25.10 & 25.10 & 25.26 \\
Ours (fine-tuned) & 30.34 & \textbf{30.32} & \textbf{30.18} & \textbf{30.09} & \textbf{30.06} & \textbf{30.20} \\
\midrule
\multirow{2}{*}{Method} & \multicolumn{6}{@{}c@{}}{Anesthesiologists being attentive} \\\cmidrule{2-7}%
 & 0.1 & 0.2 & 0.3 & 0.4 & 0.5 & Avg. \\
\midrule
ActionFormer~\citep{zhang2022actionformer}  & 18.21 & 12.37 & 8.74 & 5.28 & 3.10 & 9.54 \\
TriDet~\citep{shi2023tridet}  & 18.37 & 12.46 & 8.92 & 5.42 & 3.03 & 9.64 \\
TemporalMaxer~\citep{tang2023temporalmaxer} & 14.12 & 8.88 & 5.94 & 3.04 & 1.40 & 6.68 \\
Team-OR~\citep{chen2025they}  & 22.67 & 15.51 & 10.88 & 6.49 & 3.21 & 11.75 \\
Ours (off-the-shelf) & 28.98 & 22.91 & 17.62 & 12.59 & 7.35 & 17.89 \\
Ours (fine-tuned) & \textbf{30.95} & \textbf{24.04} & \textbf{18.40} & \textbf{13.70} & \textbf{8.31} & \textbf{19.08} \\
\bottomrule
\end{tabular*}
}
\end{table*}

\subsection{Implementation details}

We use the PyTorch framework to implement our approach with a single NVIDIA A100 GPU on the Ubuntu system. We train the gaze-following model for 15 epochs using the Adam optimizer with an initial learning rate of 1e-3. For role prediction, we train the model for 20 epochs using the AdamW optimizer with an initial learning rate of 1e-4. For phase recognition, we set $K$ to 25, and train the model for 30 epochs using the AdamW optimizer with an initial learning rate of 1e-4. For team communication detection, we train the spatial-temporal gaze model for 40 epochs with a batch size ($B$) of 1,024, and use the AdamW optimizer with an initial learning rate of 1e-4. For video preprocessing and training of the TAD model, we follow the same settings as in \citep{chen2025they}. $\tau$ in Eq.~\ref{eq:infonce_single} is set to 0.07.

\subsection{Results}
\noindent \textbf{Gaze-following in the operating room:} We compare the Gaze-LLE models pretrained on the GazeFollow dataset~\citep{recasens2015they} with the ones fine-tuned on the 4D-OR and Team-OR datasets. As shown in Table~\ref{tab:gazelle_results}, the fine-tuned models perform significantly better, indicating the necessity of domain adaptation in the OR scenes. 

\begin{table}[h]
\caption{Performance of Gaze-LLE models evaluated on the extended 4D-OR~\citep{ozsoy20224d} and Team-OR~\citep{chen2025they} datasets.}\label{tab:gazelle_results}%
\centering
\resizebox{0.35\textwidth}{!}{
\begin{tabular}{c|cc|cc}
\toprule
\multirow{2}{*}{Model} & \multicolumn{2}{c|}{4D-OR} & \multicolumn{2}{c}{Team-OR} \\
\cmidrule(lr){2-3} \cmidrule(lr){4-5}
 & AUC $\uparrow$ & L2 $\downarrow$ & AUC $\uparrow$ & L2 $\downarrow$ \\
\midrule
Pretrained (ViT-B)  & 0.9652 & 0.0938 & 0.9651 & 0.1010\\
Pretrained (ViT-L)  & 0.9697 & 0.0841 & 0.9687 & 0.0960\\
Ours (ViT-B)  & \textbf{0.9868}  & 0.0571 & 0.9898 & 0.0558 \\
Ours (ViT-L)  & 0.9860  & \textbf{0.0562} & \textbf{0.9904} & \textbf{0.0539}\\
\bottomrule
\end{tabular}
}
\end{table}



\begin{figure*}[h!]
    \centering
    \includegraphics[width=0.9\textwidth]{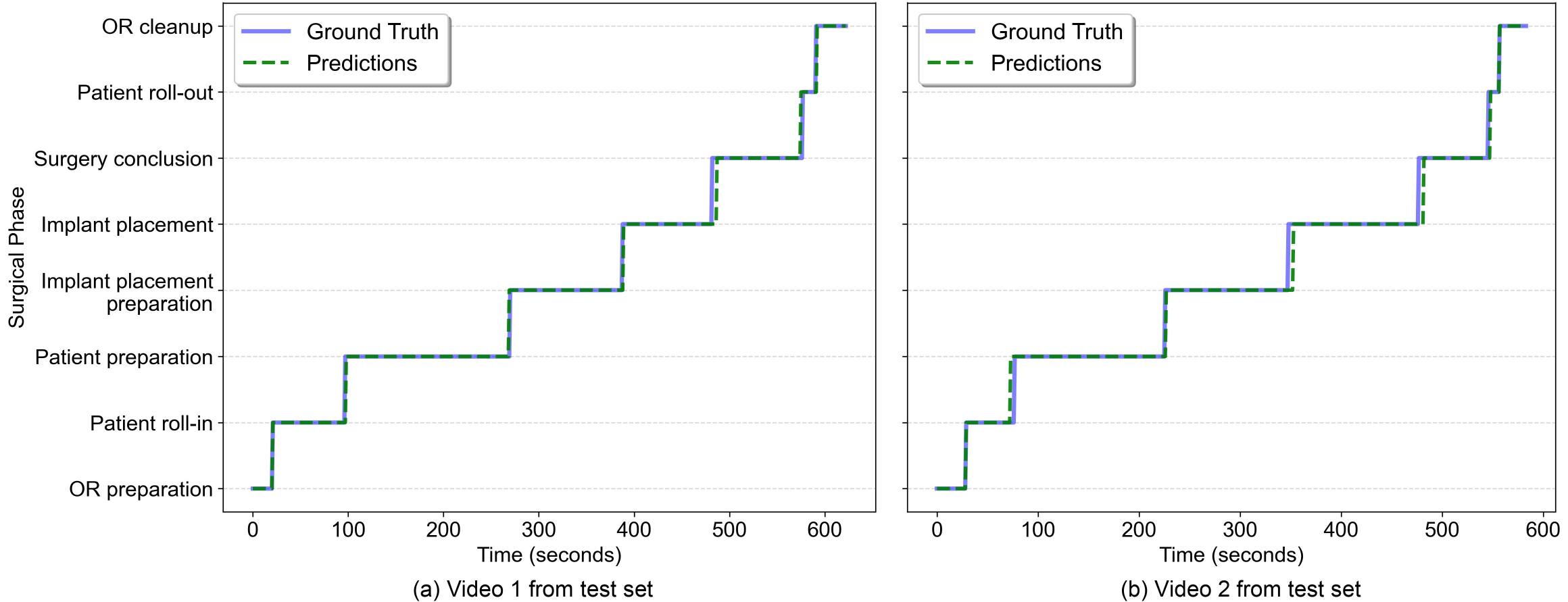} 
    \caption{Visualization of our phase predictions against the ground-truth data on 4D-OR test set.}
    \label{fig:phase}
\end{figure*}

\noindent \textbf{Clinical role prediction and surgical phase recognition:} We compare our approach with the ones that use semantic scene graph (SSG)~\citep{ozsoy2024holistic,ozsoy2023labrad,pei2024s,ozsoy2024oracle}. We train our models 10 times and report the average performance. For role prediction in Table~\ref{tab:role_prediction_comparison}, our method performs much better, showing how much the unique gaze patterns of each role contribute to the task. For phase recognition in Table~\ref{tab:phase_recognition}, our method performs slightly worse than the best ones. According to the visualization of our predictions in Fig.~\ref{fig:phase}, the errors are highly concentrated at the phase boundaries, which is understandable because we only use the ambiguous gaze predictions to recognize the phases without any other visual or semantic information. Nevertheless, our approach still achieves comparable performance compared with those that use complex SSG, which proves the significance of gaze analysis. Finally, since our approach directly uses gaze predictions as input, using a fine-tuned gaze-following model achieves much better performance than using the pretrained one. 

\begin{table}[h]
\caption{Comparison with the clinical role prediction approaches.}
\centering
\resizebox{0.45\textwidth}{!}{
\begin{tabular}{c|cccccc}
\toprule
Method & \makecell{4D-OR\\\citep{ozsoy2024holistic}} & \makecell{LABRAD\\\citep{ozsoy2023labrad}} & \makecell{$S^2$Former\\\citep{pei2024s}} & \makecell{ORacle\\\citep{ozsoy2024oracle}} & \makecell{Ours\\(off-the-shelf)} & \makecell{Ours\\(fine-tuned)} \\
\midrule
Macro F1     & 0.87 & 0.89 & 0.89 & 0.85 & 0.62 & \textbf{0.92} \\
\bottomrule
\end{tabular}
\label{tab:role_prediction_comparison}
}
\end{table}

\begin{table}[h]
\caption{Comparison with the surgical phase recognition approaches.}
\centering
\resizebox{0.45\textwidth}{!}{
\begin{tabular}{c|cccc}
\toprule
Method & 4D-OR~\citep{ozsoy2024holistic} & ORacle~\citep{ozsoy2024oracle} & Ours (off-the-shelf) & Ours (Fine-tuned) \\
\midrule
Macro F1     & 0.97 & \textbf{0.99} &  0.81 & 0.95 \\
\bottomrule
\end{tabular}
\label{tab:phase_recognition}
}
\end{table}

\begin{table*}[h]
\caption{Ablation study on detecting ``anesthesiologists being attentive''. Global and local mean we encode all the gazes and partial gazes in one frame, respectively. We report AP at different tIoU. }\label{tab:ana_abl}
\centering
\resizebox{0.8\textwidth}{!}{
\begin{tabular*}{\textwidth}{@{\extracolsep\fill}c|cccccc}
\toprule%
Method & 0.1 & 0.2 & 0.3 & 0.4 & 0.5 & Avg. \\
\midrule
Global training \& global testing & 25.54 & 19.69 & 14.98 & 11.25 & 6.55 & 15.60 \\
Local training \& local testing & 28.89 & 21.08 & 15.31 & 10.48 & 6.27 & 16.41 \\
Global training \& local testing  & \textbf{30.95} & \textbf{24.04} & \textbf{18.40} & \textbf{13.70} & \textbf{8.31} & \textbf{19.08} \\
\bottomrule
\end{tabular*}
}
\end{table*}

\noindent \textbf{Team communication detection:} We compare our approach with the state-of-the-art TAD approaches. As shown in Table~\ref{tab:team_results}, our method using both the off-the-shelf and fine-tuned gaze-following models surpasses all the existing methods. As the training of the gaze encoder is self-supervised, we can improve the performance with gaze features in any unseen OR scenes without additional labels. 

\subsection{Ablation study}\label{sec:abl}

To evaluate the effect of encoding gazes locally by heuristically only encoding the gazes of the rightmost three persons that are near the anesthesia equipment when detecting ``A.B.A.'', we conduct an ablation study with three settings: (1) encoding all gazes during training and testing; (2) encoding gazes locally during training and testing; (3) encoding all gazes during training, and encoding gazes locally during testing. As shown in Table~\ref{tab:ana_abl}, the last setting achieves the best performance.

\subsection{Limitations and future work}

In this work, we have studied gaze-following in the OR for the first time, and have proved its clinical value through gaze-based surgical workflow analysis, including role prediction, phase recognition, and team communication detection. However, there are some limitations: (1) the proposed gaze-following datasets in the OR only consist of two clinical sites, which lack the diversity for broad generalization; (2) the gaze-following research is limited to monocular 2D gaze point prediction, without studying depth information or semantic context; (3) the proposed methodology is fully-supervised, which requires ground-truth annotations; (4) the proposed heuristic strategy of encoding gazes of the rightmost three persons for team communication detection disables generalizability to the other OR scenes. In the future, we plan to further expand gaze research in the OR in several directions: (1) 3D semantic gaze-following, where we can localize and classify the gaze target in 3D space; (2) unsupervised domain adaptation, where we can follow the gaze in any new clinical sites without additional annotations; (3) gaze-assisted semantic scene graph generation, where we can predict human-human and human-object interactions more accurately through gaze analysis; (4) few-shot learning, where we can recognize roles, phases, and activities given a few examples.  

\section{Conclusion}\label{sec:conclusion}

In this paper, we introduce a novel perspective for surgical workflow analysis through gaze-following, which infers where clinicians are looking in the operating room. In particular, we extend the 4D-OR and Team-OR datasets with gaze annotations, create annotations of a new activity class in Team-OR, and use gaze predictions to address clinical role prediction, surgical phase recognition, and team communication detection. Experimental results prove the value of gaze analysis, and we envision that gaze-following can be further explored to benefit surgical data science and computer-assisted interventions. 

\section{Acknowledgments}

This work was supported by French state funds managed within the Plan Investissements d’Avenir by the ANR under references ANR-22-FAI1-0001 (project DAIOR), ANR-10-IAHU-02 (IHU Strasbourg) and by BPI France (Project 5G-OR). This work was also supported by Ruth \& Arthur Scherbarth Foundation.
This work was granted access to the servers/HPC resources managed by CAMMA, IHU Strasbourg, Unistra Mesocentre, and GENCI-IDRIS [AD011014722R2, AD011011631R4, and AD011011638R3]. We further thank Andreas Weibel and Lionel Bergerot for setting up the recording system in the operating room and for providing technical support throughout the study, and Laurin Terhorst for assisting with data collection.
\\

\textbf{Ethical approval:} The project is conducted under the following protocols: the Swiss legal requirements, the current version of the World Medical Association Declaration of Helsinki, the principles and procedures for integrity in scientific research involving human beings, the {\it StOP? II} clinical trial and the study ``Assessment of team coordination in the operating room based on motion analysis: A proof of concepts study''.

\textbf{Competing interests:} The authors declare no conflict of interest. 

\textbf{Informed consent:} Data was collected with informed consent from the human participants involved.

\bibliographystyle{model2-names.bst}
\bibliography{sn-bibliography}

\end{document}